\documentclass[a4paper,fleqn]{cas-dc}

\usepackage[numbers]{natbib}

\usepackage{graphicx}
\usepackage{caption}
\usepackage{subcaption}
\usepackage{booktabs}
\usepackage{amssymb}
\usepackage{amsmath}
\usepackage{enumitem}
\usepackage{paralist}
\usepackage{float}
\usepackage{algorithm}
\usepackage{algorithmic}
\usepackage[section]{placeins}
\usepackage{appendix}

\newif\ifyshrink

\yshrinkfalse

\author[1,2]{Hongxiao Li}
\ead{lihongxiao19@mails.ucas.ac.cn}

\author[1,2]{Wanling Gao}
\ead{gaowanling@ict.ac.cn}

\address[1]{Institute of Computing Technology, Chinese Academy of Sciences, Beijing, China}
\address[2]{University of Chinese Academy of Sciences, Beijing, China}
\begin{document}

\let\WriteBookmarks\relax
\def\floatpagepagefraction{1}
\def\textpagefraction{.001}

\title [mode = title]{AutoMatBench: An Automatic Optimization Toolkit for the Acceleration of Material Properties Prediction Benchmarking}                      

\shorttitle{AutoMatBench: An Automatic Optimization Toolkit}

\begin{abstract}
Material property prediction (MPP) infers key properties from chemical composition and structure, accelerating the discovery and optimization of novel materials. In the realm of MPP, MatBench is a widely accepted benchmarking tool that defines over ten significant problems and provides the paradigm of performance evaluation for AI prediction models. Even though MatBench works well in benchmarking the performances of prediction models on in-distribution (ID) tasks and datasets, it lacks the ability to reflect their performances on out-of-distribution (OOD) material data, resulting failure in new material discovery. By combining the pipelines of MatBench and the existing researches on OOD performance evaluation, this study enables a huge space of benchmarking configurations, comprehensively reflecting the performances, abilities, and disadvantages of various AI prediction models. This work reports that the discrepancy of performances at different configuration values is huge and can be illustrated with prior knowledge and novel insights, therefore consideration of causal effect of configurations on performance results is necessary. In case of the impossibility of enumerative benchmarking at every configuration, this work further proposes AutoMatBench, an automatic toolkit with Bayesian optimization. Experiments with AutoMatBench reports that, within twelve steps of optimization, the similar results with MatBench and former OOD research can be accessed while more than half of the cost are saved. Besides, this tool also yields more essential findings on MPP benchmarking, positively contributing to the cost and efficiency of new material discovery.
\end{abstract}


\maketitle

\section{Introduction}

Material property prediction (MPP), as a signficant sub-field of Artificial Intelligence for Science (AI4S), aims at predicting or inferring the physical, chemical, or functional properties of materials directly from their composition and structure data. By learning relationships of compositional or structural information from experimental and computational data, MPP models enable high-efficiency screening of demanded materials, guide target-oriented synthesis, and reduces reliance on high-cost experiments or \textit{ab initio} simulations.

MatBench~\cite{dunn2020benchmarking} emerges as a widely accepted benchmarking suite that defines over ten diverse representative prediction tasks for both classification and regression problems, curated from solid datasets. MatBench provides a consistent training-test splitting and cross-fold validation, establishing a generalized and comparable paradigm for evaluating the prediction accuracies of AI models across tasks and datasets. It has become the most adopted \textit{de facto} reference for developing and validating new material informatics methods.

Unfortunately, benchmarking results with MatBench only represent the performances on in-distribution (ID) tasks and datasets, while the performances on out-of-distribution (OOD) data are totally different. Research from Omee et al.~\cite{omee2024structure} and Fung et al.~\cite{fung2021benchmarking} also support such conclusion. Experiments of this work also report similar outcomes.

This study aims at establishing an evaluation methodology the abilities of AI prediction models that works both on ID and OOD data, referring the pipelines of MatBench~\cite{dunn2020benchmarking} and the existing OOD performance evaluation. There are two main challenges. On the one hand, the existing OOD performance evaluation~\cite{segal2025known,tan2025benchmarking,li2025out,omee2024structure} are neither completely undergone under the same or similar paradigm of MatBench, nor including the majority of material prediction tasks that MatBench owns. On the other hand, the results of both MatBench and OOD research relies on the specific benchmarking configurations of experiments, which fluctuate heavily when changed. As a result, asserting which performance result is the actual ability of an AI model is non-trivial.

To solve such problems, we basically adopt the workflow of Omee et al.~\cite{omee2024structure} and the dataset source of MatBench~\cite{dunn2020benchmarking}, but with a larger space of hundreds of common configurations and a larger data scope. We select six tasks: band gap~\cite{jain2013commentary,ong2015materials,zhuo2018predicting}, bulk modulus~\cite{jain2013commentary,ong2015materials,de2015charting}, shear modulus~\cite{jain2013commentary,ong2015materials,de2015charting}, formation energy~\cite{jain2013commentary,ong2015materials,castelli2012new}, refractive index~\cite{jain2013commentary,ong2015materials,petousis2017high}, and metallicity~\cite{zhuo2018predicting}, which can be aligned to MatBench~\cite{dunn2020benchmarking} and is the superset of Omee et al.~\cite{omee2024structure}'s research scope.

By evaluation of five general-purpose prediction models that can be aligned to the officially announced results of MatBench~\cite{dunn2020benchmarking}: ALIGNN~\cite{choudhary2021atomistic}, SchNet~\cite{schutt2017schnet,schutt2018schnet}, CrabNet~\cite{wang2021compositionally}, RF-SCM~\cite{dunn2020benchmarking,kovacs2021lsst,faber2015crystal,ward2016general}, and MEGNet~\cite{chen2019graph}, we report the discrepancy of performances at different configuration values, where configuration refers to three factors: dataset ratio $r$, clustering number of OOD material data $n$, and the number of centroid and nearest neighbor $s$, which are the same semantics as in those of OOD research. Our results report that the performance for a specific model on a certain task has a huge discrepancy between different configurations. Merely deciding the result of any single configuration or a few configurations as the ability of the model may involve heavy biases and misunderstanding of the model.

This issue splits into two occasions. First, if the ground-truth performance exists, how to find the closest configuration at low cost; second, if not (or cannot be well-defined), which set of configurations still makes sense in benchmarking the models. For the first problem, our answer is developing an automatic optimization toolkit, AutoMatBench; for the second one, our aim is to search for the configurations that discriminate performances of different models.

We propose AutoMatBench, an automatic toolkit with Bayesian optimization~\cite{frazier2018tutorial}. The experiment results report that, within twelve steps of optimization, AutoMatBench manages to find the configuration (i.e., aforementioned $(r,n,s)$ tuples) that meets the goal for all tasks and AI models, in accordance of the existing research outcome, but saves more than half of the cost. When the steps are set to more than twelve, improvement is limited. Moreover, AutoMatBench also delivers more findings for MPP benchmarking that are not reported in the existing studies, and can be well-illustrated with prior knowledge in MPP field and novel findings of this work.

We hope the study of AutoMatBench guide the selection of high value-for-money algorithmic configurations, reduce experimental and computational costs, and support end-to-end optimization from method selection and data governance to deployment, informing model design and feature engineering for not only MPP but also other AI4S issues.

\section{Background and related work}

\subsection{Background}

This study focuses on the evaluation and automatization in materials informatics field. At the foundation of materials informatics lie \textit{ab initio} calculations, most notably density functional theory (DFT) (e.g., DFT-GGA~\cite{perdew1996generalized}, DFT-vDW-DF~\cite{dion2004van}), the gold standard for generating accurate, atomic-level materials property data with well-defined physical consistency started from physical theorems.

The most prominent material datasets are the Materials Project (MP)~\cite{jain2013commentary} (over 154,000 inorganic materials with DFT-calculated thermodynamic, electronic, and mechanical properties), Open Quantum Materials Database (OQMD)~\cite{saal2013materials} (crystalline materials' structural and energetic data), and the JARVIS infrastructure~\cite{choudhary2020joint} (over 40,000 materials' standardized DFT datasets and integrated machine-learning (ML) tools). These databases have democratized high-quality material data access and established a unified foundation for property prediction ML model training and validation.

Our proposal of AutoMatBench is based on MatBench~\cite{dunn2020benchmarking} -- analogous to ImageNet~\cite{deng2009imagenet} -- offering thirteen standardized classification and regression tasks with DFT and experimental data, covering a variety of materials, serving as the \textit{de facto} benchmarking suite for validating materials ML models like graph neural networks (GNNs).

Although SOTA ML models perform well on ID benchmarks like MatBench~\cite{dunn2020benchmarking}, real-world materials discovery requires generalization to OOD samples. Traditional random splitting strategies overestimate performance and ignore real extrapolation challenges. Omee et al.~\cite{omee2024structure} first systematically benchmarked GNNs on OOD materials property prediction with five realistic tasks. They found a large generalization gap that most top MatBench models degraded severely under OOD settings, while only a few (e.g., ALIGNN~\cite{choudhary2021atomistic}) remained relatively robust. This work revealed the weakness of conventional ID benchmarks and spurred research on OOD-robust methods for materials informatics. The framework of AutoMatBench follows the majority of principles of Omee et al.~\cite{omee2024structure}.

Beyond domain-specific benchmarks, Evaluatology~\cite{zhan2025evaluatology} has been established as a universal theoretical framework for rigorous, consistent evaluation across disciplines, addressing fundamental challenges in evaluation. It focuses on distinguishing the causal effect of a particular object from the total effect of an entire system. The problem definition of our work is in accordance with Evalautology~\cite{zhan2025evaluatology}.

Our study also involves automatic optimization of benchmarking techniques. Existing optimization methods include gradient-based optimizers (i.e., SGD~\cite{amari1993backpropagation}, Adam~\cite{kingma2014adam}), non-gradient algorithms, and model compression techniques. We only adopt Bayesian optimization~\cite{frazier2018tutorial}, a sample-efficient black-box optimization approach that uses a surrogate model and acquisition function to balance exploration and exploitation.

\subsection{Other related work}

Besides, there are other studies of OOD materials. Segal et al.~\cite{segal2025known} specifically addresses the issue of OOD attribute value prediction (Y-extrapolation) by introducing transduction learning methods. Tan et al. proposes MatUQ~\cite{tan2025benchmarking}, aiming to jointly evaluate the prediction accuracy and Uncertainty Quantification (UQ) quality of GNNS in OOD scenarios, emphasizing the need for prediction reliability in actual material discovery. Li et al. proposes the Crystal Adversarial Learning (CAL) algorithm~\cite{li2025out}, specifically for the small sample problem.

Before our research, there is the study of SAIBench~\cite{li2022saibench,li2023saibench}, which is based on Evaluatology~\cite{zhan2025evaluatology} and the physical symmetricity of AI4S tasks, an AI4S benchmark suite that includes the tasks of machine learning for force field (MLFF), jet tagging, and precipitation nowcasting, provides a unified benchmarking system for scientific AI, decoupling scientific research problems, AI models, evaluation metrics, and computational configurations into reusable modules to standardize benchmarking across diverse scientific domains.

\section{Motivation}\label{sec:motivation}

\begin{figure}[ht]
\centering
\includegraphics[scale=0.3]{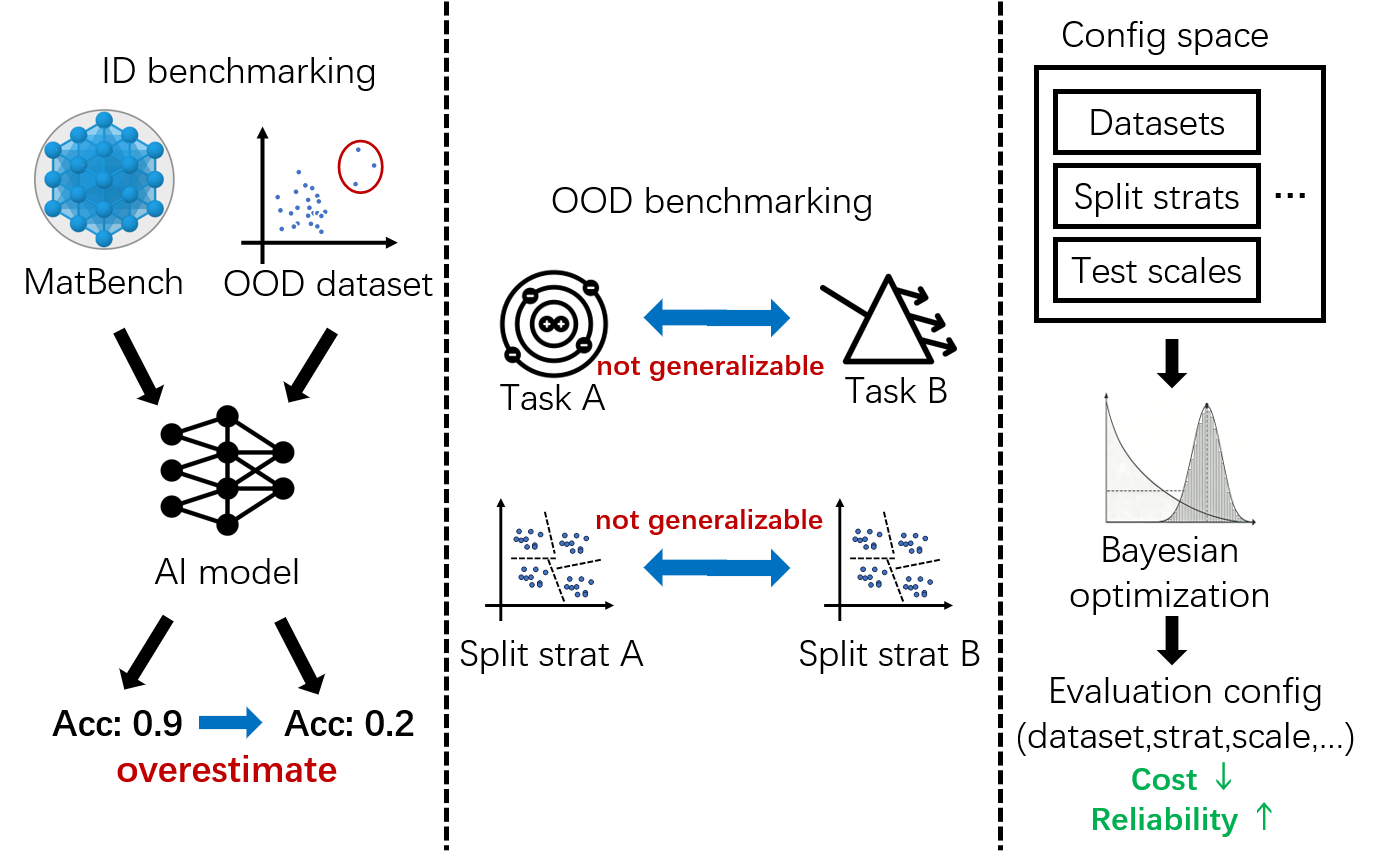}
\caption{The motivation: this study provide a novel, cost-effective, and near-reliable benchmarking method for material that solves both problems of failure in OOD evaluation and the lack of generalizability.}\label{fig:motivation}
\end{figure}

Figure~\ref{fig:motivation} presents the overall motivation of this work.

\subsection{MatBench has limited ability on benchmarking the OOD performance of ML models}

MatBench~\cite{dunn2020benchmarking} evaluative capacity is predominantly restricted to ID scenarios. According to its dataset split strategy, the test dataset adheres to the same statistical distribution as the training dataset. In practical materials science research and applications, ML models frequently encounter OOD challenges, that is, predicting the properties of materials whose compositions, crystal structures, or properties (Y-values) are distant from the distribution center of the training dataset. Omee et al.~\cite{omee2024structure} reports that all GNN models for material property prediction exhibit a substantial performance gap between OOD and ID tasks: their OOD MAEs are significantly higher than ID baselines with degradation ranging from $-0.83\%$ to $-5568.31\%$. Only CGCNN~\cite{xie2018crystal}, ALIGNN~\cite{choudhary2021atomistic} and DeeperGATGNN~\cite{omee2022scalable} achieve slight ID performance surpassing on a few OOD targets. This inherent limitation tends to result in severe overestimation of model performance in real-world applications, thereby impeding the efficiency of AI models in materials discovery and rational design.

\subsection{The existing OOD benchmarking methods has a huge cost and poor generalizability}

To address the aforementioned gap in OOD evaluation, several novel models and OOD benchmarking approaches have been proposed in recent years~\cite{choudhary2021atomistic,omee2022scalable,chen2019graph,de2021materials}. However, these methods are plagued by two critical limitations: high implementation cost and poor generalizability. On the one hand, due to the high cost of OOD material discovery from the laboratories, only limited scale if OOD data are obtained. For example, the MP~\cite{jain2013commentary} and JARVIS~\cite{choudhary2020joint} dataset include the majority of known material data. However, the OOD data ratio is around $15\%$ or $20\%$. Therefore, conducting OOD evaluations across multiple model architectures and task types demands considerable computational resources, as it involves retraining models on diverse data splits and performing repeated evaluative experiments. On the other hand, the majority of existing OOD benchmarking methods are task-specific -- they are designed for a particular materials property (e.g., band gap, formation energy, etc.), which renders them difficult to extend to other materials-related tasks or OOD settings. This lack of generalizability significantly limits their utility as a universal evaluative tool for assessing OOD performance across the broader material ML community. According to our estimation, the cost of a complete evaluation on six tasks is up to orders of magnitude than benchmarking with MatBench~\cite{dunn2020benchmarking}.

\subsection{Automatic optimization finds cost-effective evaluation configurations with near-reliable results}

Given the inherent limitations of MatBench~\cite{dunn2020benchmarking} and the critical shortcomings of existing OOD benchmarking methods, there exists an urgent demand for a cost-effective and generalizable approach to OOD evaluation in materials ML. Automatic optimization offers a promising solution to this pressing challenge. By leveraging an optimization algorithm, it is feasible to automatically search for optimal OOD evaluation configurations -- including material dataset ratio, data splitting strategy, and OOD data scale decided by the number of nearest neighbor by clustering -- that strike a balance between evaluation cost and result reliability. This study focuses on three concrete problems. First, whether the evaluated performances of different configurations vary, which influences the reliability of a single evaluation. Second, if there is a ground-truth performance of a certain AI model, how to find the evaluation configuration that yields the closest result at low lost. Third, without defining the ground-truth performance, what configurations discriminates most AI models' performances. We managed to formalize, solve, and implement all of the above with the Bayesian optimization algorithm~\cite{frazier2018tutorial} as AutoMatBench. The experiment results of this research reports that within twelve steps of optimization, the adequate configurations of each problem are found (even though not the real optimum). This study further finds better configurations with less than half ($<50\%$) of the cost compared to Omee et al.'s method~\cite{omee2024structure}.

\section{Problem definition}\label{sec:problem}

\subsection{Evaluation configuration space}

This study focuses on the performances of material prediction models on certain material datasets. All the prediction tasks in this research are classification problems. The index of models' performance is chosen as the mean average error (MAE). Section~\ref{sec:framework} in late text provides detailed description of OOD benchmarking pipeline. Briefly, sampling from two datasets, material clustering into groups, and retrieving nearest neighbors are three critical steps that includes parameters.

For formalization, we note the total set of AI models as $O$ and a certain model as $o_i\in O$ ($i$ for a certain model). The ratio $r$ denotes the ratio of material data sampled from JARVIS dataset~\cite{choudhary2020joint}, while the rest are sampled from MP dataset~\cite{jain2013commentary}. The letter $n$ denotes the number of folds in clustering data with K-Means algorithm~\cite{ahmed2020k}. The letter $s$ denotes the number of nearest neighbors of every cluster centroid plus the centroid itself. The details are provided in Section~\ref{sec:framework}. We use capital letters $R$, $N$, and $S$ to represent the value range of $r$, $n$, and $s$, which includes multiple choices, provided in Section~\ref{sec:framework}. We note the Cartesian product of them as the evaluation configuration space $C=R\times N\times S$, which includes every probable setting in our work. This definition enables to evaluate an AI model with any dataset generated with a given configuration $c_j=(r,n,s)$ ($j$ for a certain choice), which is the basis of the following analysis. The performance of a model $o_i$ on a certain dataset $c_j$ is noted as $m(o_i,c_j)$.

This work studies the following problems: firstly, quantify the disparity between models' performances with different configuration; secondly, find the best configuration at low cost when ground-truth exists; lastly, choose a rational configuration when ground-truth is not defined. They are formalized as follows respectively.

\subsection{The first-class evaluation problem}

This problem is the most common consideration of material science researchers. It aims at finding the model of best average performance on a certain material dataset. Formally, we have:
\begin{equation}\label{for:1}
    \arg\max_{o_i\in O}\; \operatorname*{mean}_{c_j\in C} m(o_i,c_j)
\end{equation}
Especially, if top-K performance rather than the average is considered, the $\operatorname*{mean}$ operator can be changed to $\operatorname*{top}_K$ correspondingly. This work studies the stability of this value when the evaluation configuration is changed.

\subsection{The dual of first-class evaluation problem}

This problem is the mathematical dual of the above problem, if the $\operatorname*{mean}$ operator is changed to $\operatorname*{range}$. It aims at solving such problem: among all the configurations, which one has the maximum quantity discrimination ability among AI models, where ``quantity'' refers to its performance (MAE).
\begin{equation}\label{for:1dual}
    \arg\max\limits_{c_j\in C}\operatorname*{range}\limits_{o_i\in O}m(o_i,c_j)
\end{equation}
A dataset with the better discrimination ability on models is more acceptable.

\subsection{The second-class evaluation problem}

This problem is another essential problem that focuses on the validity and cost of benchmarking. In this context, we suppose the ground-truth quantity of an AI model is formally definite. It can refer to a certain reliable benchmarking research or a conventionally consensus of the AI4S community. Otherwise it can be defined as the aggregation (average, top-K, or other representations) of AI performances in the entire configuration space $C$ as follows, according to the concrete research requirement:
\begin{equation}\label{for:gt}
    \mathrm{gt}(o_i,C)\stackrel{\mathrm{def}}{=}\operatorname*{agg}\limits_{c_j\in C}(o,c_j)
\end{equation}
Based on the definition of $\mathrm{gt}$, the problem is defined as finding the best configuration set that yields results close to the $\mathrm{gt}$ value at low cost:
\begin{equation}\label{for:2}
    \arg\min\limits_{o_i\in O}\operatorname*{mean}\limits_{c_j\in C}\mathrm{dist}(m(o_i,c_j),\mathrm{gt}(o_i,C))
\end{equation}
A result that has a huge disparity with the ground-truth result is considered as invalid or flawed. 

\section{Methodology}

\subsection{The existing methods}

Based on the problem definitions, the MatBench~\cite{dunn2020benchmarking}'s benchmarking problem can be reformed as follows:
\begin{equation}\label{for:existing}
    \arg\max\limits_{o_i\in O}m(o_i,c_0)
\end{equation}
where $c_0$ is a certain dataset for a given task, and different $o_i$'s are AI models to be evaluated.

For example, suppose the task is the band gap prediction problem~\cite{jain2013commentary,ong2015materials,zhuo2018predicting}, then the dataset $c_0$ is derived from Zhuo et al.~\cite{zhuo2018predicting}. According to the rules defined by MatBench~\cite{dunn2020benchmarking}, the dataset is separated into several folds that cover training set and test set, following the Nested CV scheme~\cite{wainer2021nested}. An $o_i$ may refer to a certain model (e.g., ALIGNN~\cite{choudhary2021atomistic}).

For OOD benchmarking, the method of Omee et al.~\cite{omee2024structure} is similar but for several differences. First, the dataset includes only screened OOD materials, rather than the entire primitive dataset. Second, the training-test set splitting is in accordance with the clustering of materials, that is, every fold belongs to a certain cluster, representing materials that have similar characteristics or properties. The formalized problem remains unchanged.

\subsection{Our novelty}

As defined in Section~\ref{sec:problem}, our research focuses on three major problems different from the existing. According to Section~\ref{sec:motivation}, the performances of models vary to orders of magnitude from ID problems to OOD problems. In addition, using only a mere fixed OOD dataset, the results with different configurations $(r,n,s)$ fluctuates heavily. Therefore, this study does not only consider the performance results on any fixed dataset with its separation.

We first define the total dataset that contains the majority of important ID and OOD material data. The ideal goal is to evaluate every model's performance on the entire dataset. Unfortunately, it is impossible for cost concerns. As a substitute, we evaluate on different samples with the dataset to approximate the ground-truth performances of models.

\begin{figure}[ht]
\centering
\includegraphics[scale=0.4]{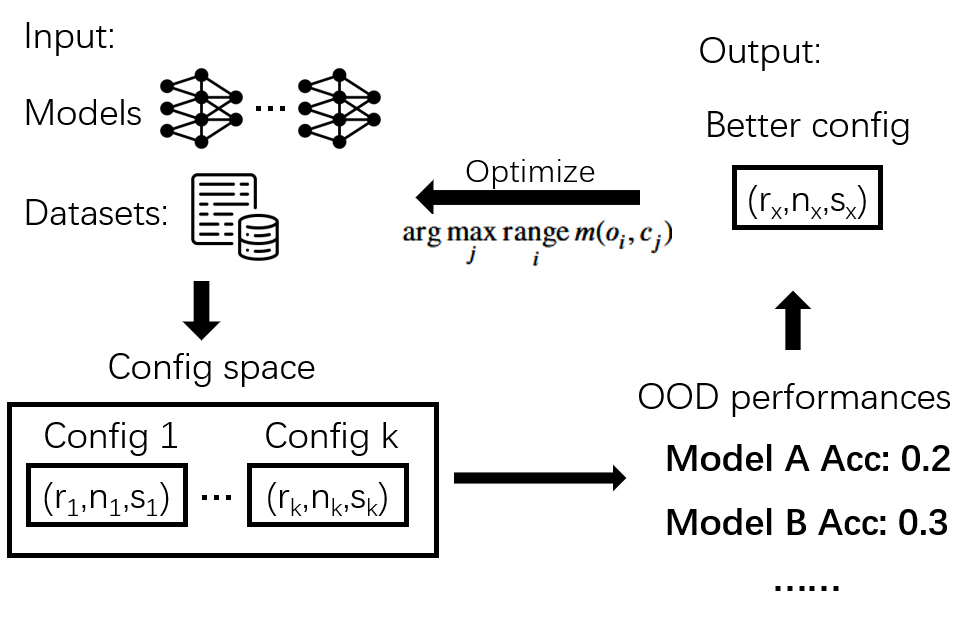}
\caption{The methodology of this research. The other existing benchmarking methods do not include configuration adjustment and optimization parts. The optimization part is the essence of our method.}\label{fig:methodology}
\end{figure}

Two main challenges exist with this approach. First, how the ground-truth performance is defined. This issue is explained in the second-class problem in late text. Second, what algorithm approximates the ground-truth performance at low cost. For all the three problems, the Bayesian optimizer~\cite{frazier2018tutorial} works well. Figure~\ref{fig:methodology} is the overall view of our methodology.

\subsection{The algorithm for the first-class problem}

To address the first-class evaluation optimization problem under a limited evaluation budget, we adopt Bayesian optimizer~\cite{frazier2018tutorial} to efficiently identify the optimal AI model that maximizes average performance (MAE) across all evaluation configurations. The core objective is to obtain the optimal model $\hat{o}$ by maximizing the mean evaluation metric over the full configuration set, defined with formula~\ref{for:1} where $o$ denotes a candidate AI model, $c$ denotes an evaluation configuration, and $m(o,c)$ measures the performance of model $o$ under configuration $c$. Exhaustive evaluation of all model-configuration pairs is computationally infeasible, making Bayesian optimization an ideal sample-efficient solution.

Algorithm~\ref{alg:1} operates iteratively within a fixed evaluation budget $T$, split into two sequential phases: initial random exploration and adaptive optimization. It first initializes the dataset with a small set of random $(o,c)$ samples to exhaust an initial portion of the budget, prioritizing broad exploration of unobserved $(o,c)$ pairs to map the latent performance landscape. Once the random exploration phase is complete, this procedure iteratively updates a probabilistic surrogate model using the collected data and leverages an acquisition function to select the most informative $(o,c)$ pair for evaluation -- this optimization phase balances targeted exploration of under-sampled regions and exploitation of high-performance regions to maximize evaluation efficiency. After fully exhausting the total evaluation budget, we estimate the average performance $\hat{M}(o)$ for each AI model across all evaluated configurations, and select the model with the highest estimated average performance as the optimal solution. This algorithm avoids redundant evaluations, adapts dynamically to the underlying performance landscape, and reliably locates the optimal AI model under strict resource constraints.

\begin{algorithm}[t]
\caption{Bayesian optimization for the first-class evaluation problem}
\begin{algorithmic}[1]\label{alg:1}
\REQUIRE 
    Option set $O = \{o_1, o_2, \cdots\}$,\\
    Context set $C = \{c_1, c_2, \cdots\}$,\\
    Evaluation function $m(o,c)$,\\
    Evaluation budget $T$,\\
    Initial sample size $T_0$.
\ENSURE 
    Optimal option $\hat{o} \in O$ that maximizes $\widehat{M}(o)=\frac{1}{|C|}\sum_{c_j \in C} m(o,c_j)$
\STATE Initialize dataset $D \leftarrow \emptyset$
\STATE Random sample and evaluate $T_0$ $(o,c)$ pairs into $D$
\FOR{$t=1$ \TO $T-T_0$}
    \STATE Update surrogate model with $D$
    \STATE Select $(o^*,c^*)$ by acquisition function
    \STATE Evaluate and update dataset $D$
\ENDFOR
\STATE Estimate $\widehat{M}(o)$ for all $o \in O$
\STATE Return $\hat{o} = \arg\max_o \widehat{M}(o)$
\end{algorithmic}
\end{algorithm}

\subsection{The algorithm for the dual of first-class problem}

Similar to the first-class problem, its dual problem aims to obtain the optimal configuration $\hat{c}$ by maximizing the range evaluation metric over the full configuration set, defined with formula~\ref{for:1dual}. After fully exhausting the total evaluation budget, we estimate the range of models' performances $\hat{R}(c)$ under the configuration $c$ instead for each AI model across all evaluated configurations, and select the configuration with the maximum estimated range as the optimal solution. The corresponding algorithm is Algorithm~\ref{alg:1dual}.

\begin{algorithm}[t]
\caption{Bayesian optimization for the dual of first-class evaluation problem}
\begin{algorithmic}[1]\label{alg:1dual}
\REQUIRE 
    Option set $O = \{o_1, o_2, \cdots\}$,\\
    Context set $C = \{c_1, c_2, \cdots\}$,\\
    Evaluation function $m(o,c)$,\\
    Evaluation budget $T$,\\
    Initial sample size $T_0$.
\ENSURE 
    Optimal context $\hat{c} \in C$ that maximizes $\widehat{R}(c)=\min\limits_{o_{i_1},o_{i_2} \in O} |m(o_{i_1},c)-m(o_{i_2},c)|$
\STATE Initialize dataset $D \leftarrow \emptyset$
\STATE Random sample and evaluate $T_0$ $(o,c)$ pairs into $D$
\FOR{$t=1$ \TO $T-T_0$}
    \STATE Update surrogate model with $D$
    \STATE Select $(o^*,c^*)$ by acquisition function
    \STATE Evaluate and update dataset $D$
\ENDFOR
\STATE Estimate $\widehat{R}(c)$ for all $c \in C$
\STATE Return $\hat{c} = \arg\max_c \widehat{R}(c)$
\end{algorithmic}
\end{algorithm}

\subsection{The algorithm for the second-class problem}

For the second-class problem, the target is to obtain the optimal configuration $\hat{c}$ by minimizing the disparity of performances between its result and the ground-truth over the full configuration set, defined with formula~\ref{for:2}. The ground-truth function $\mathrm{gt}(o,C)$ denotes the ground-truth performance value of option $o$ measured on the complete context set $C$, according to Formula~\ref{for:gt}. We estimate and select the configuration with the minimum distance to the ground-truth as the optimal solution. The corresponding algorithm is Algorithm~\ref{alg:2}.

\begin{algorithm}[t]
\caption{Bayesian optimization for the second-class evaluation problem}
\begin{algorithmic}[1]\label{alg:2}
\REQUIRE 
    Option set $O = \{o_1, o_2, \cdots\}$,\\
    Context set $C = \{c_1, c_2, \cdots\}$,\\
    Evaluation function $m(o,c)$,\\
    Ground-truth function $\mathrm{gt}(o,C)$,\\
    Distance function $\mathrm{dist}(\cdot,\cdot)$,\\
    Ground-truth $\mathrm{gt}(o_i,C)=\operatorname*{agg}\limits_{c_j\in C}(o,c_j)$,\\
    Evaluation budget $T$,\\
    Initial sample size $T_0$.
\ENSURE 
    Optimal option $\hat{o} \in O$ that minimizes $\widehat{L}(o)=\frac{1}{|C|}\sum_{c_j \in C} \mathrm{dist}\big(m(o,c_j),\mathrm{gt}(o,C)\big)$
\STATE Initialize dataset $D \leftarrow \emptyset$
\STATE Random sample and evaluate $T_0$ $(o,c)$ pairs into $D$
\FOR{$t=1$ \TO $T-T_0$}
    \STATE Update surrogate model with $D$
    \STATE Select $(o^*,c^*)$ by acquisition function
    \STATE Evaluate and update dataset $D$
\ENDFOR
\STATE Estimate $\widehat{L}(o)$ for all $o \in O$
\STATE Return $\hat{o} = \arg\min_o \widehat{L}(o)$
\end{algorithmic}
\end{algorithm}

\section{Framework}\label{sec:framework}

\subsection{The evaluation pipeline}

\textbf{Step 1 (Sampling)}: Sample $n_{total}=1,600$ materials from the MP~\cite{jain2013commentary} and JARVIS~\cite{choudhary2020joint} databases according to a prescribed proportional ratio $r$. For every task, all material data from MP~\cite{jain2013commentary} and JARVIS~\cite{choudhary2020joint} that include the corresponding property values are downloaded. After that, $n_{total}$ data are randomly sampled from the database. The savefile of every material includes the original CIF file (generated using \texttt{pymatgen} or \texttt{JARVIS-Tools}), the $n_{dim}=1,024$ dimensional OFM characteristic, and the property (Y-value).

\textbf{Step 2 (Dimensionality reduction)}: Embed the high-dimensional structural descriptors into a 2D space using t-distributed stochastic neighbor embedding (t-SNE, using\\\texttt{sklearn.manifold.TSNE})~\cite{van2008visualizing}. This step preserves local neighborhood structure to reveal clusters and similarities in the high-dimensional data, making latent structural patterns more readily identifiable in the 2D visualization.

\textbf{Step 3 (Density estimation)}: Estimate local sample densities on the 2D embedding via Gaussian kernel density estimation (KDE, using \texttt{scipy.stats.gaussian$\_$kde})~\cite{parzen1962estimation}. This step approximates the continuous probability density of points on the 2D manifold, highlighting high-density regions (clusters) and low-density gaps by smoothing with a Gaussian kernel.

\textbf{Step 4 (Sparse screening)}: Select the $n_{sample}=480$ samples with the lowest KDE values and designate them as out-of-distribution (OOD) candidates. The selected candidates that have the lowest density represent the materials of 2D OOD characteristics.

\textbf{Step 5 (Sparse clustering)}: Cluster the $n_{sample}$ OOD candidates into $n$ groups using K-means ($n$-fold)~\cite{ahmed2020k}. This step partitions the $n_{sample}$ OOD candidates into $n$ compact, non-overlapping clusters by minimizing within-cluster variance, yielding representative groups for next-step analysis.

\textbf{Step 6 (Target generation and split)}: For each cluster, choose its centroid as an anchor and, in the original high-dimensional space, retrieve the $s$ nearest neighbors of the corresponding sample; the union of these neighbors forms the test set, and the remaining samples constitute the training set. For clarity, if $s=1$, then no nearest neighbors are taken. This measure corresponds to the SparseX-Single strategy in Omee et al.'s benchmark~\cite{omee2024structure}. If $s>1$, then there will be $s-1$ nearest neighbors, corresponding to the SparseX-Cluster strategy instead. The final $n$-fold data for cross-validation is as follows: each fold corresponds to a certain cluster, representing an unknown or known type of materials. In each fold, a proportion $p=0.1$ of data are considered as the validation set. For the rest, the test set includes only $s$ data. The training set consists of all samples except this cluster.

\subsection{The optimization settings}

\begin{figure}[ht]
\centering
\includegraphics[scale=0.4]{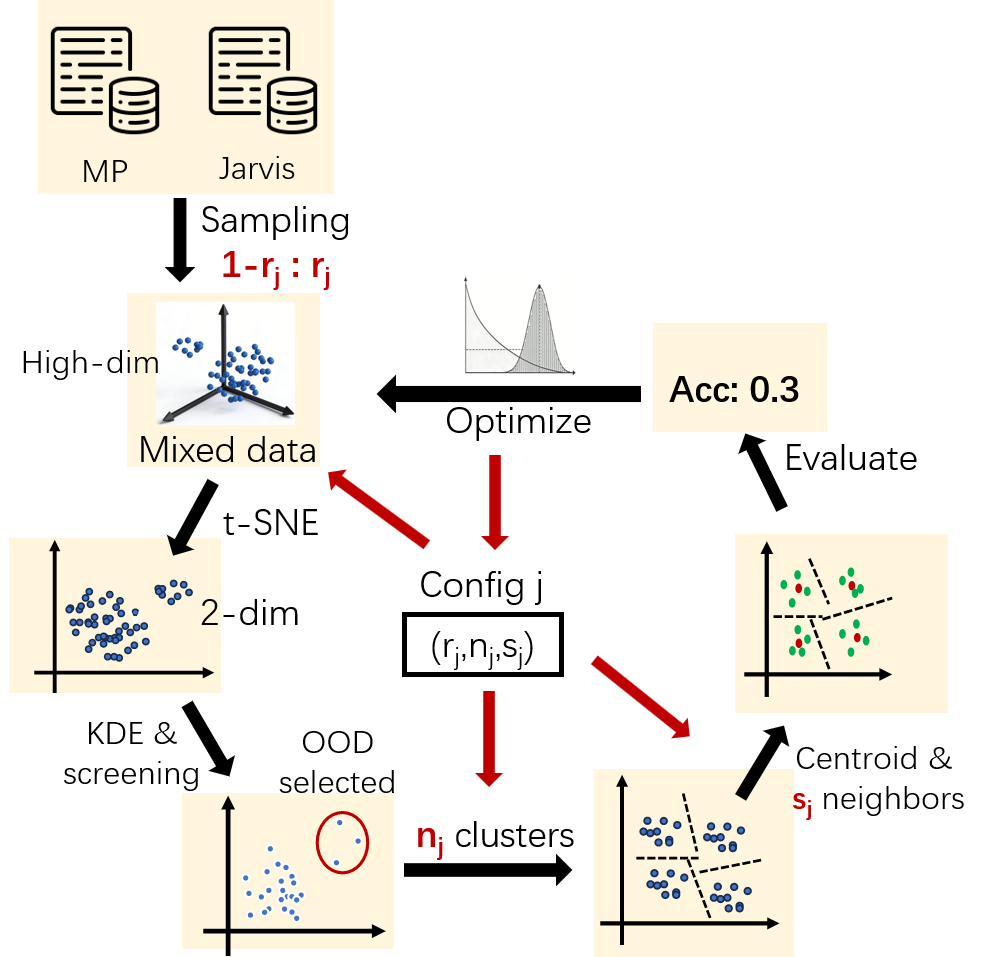}
\caption{This figure presents the framework of AutoMatBench that consists if several modules. The pipeline is represented with procedures with arrows.}\label{fig:pipeline}
\end{figure}

Figure~\ref{fig:pipeline} presents the entire procedure. The cycle on the bottom describes a one-step evaluation under the configuration $(r,n,s)$. As for optimization algorithms (\ref{alg:1},\ref{alg:1dual}, and \ref{alg:2}), the evaluation budget is set as $T=12$, and the exploration budget of the initial dataset $D$ is set as $T_0=4$. To ensure a fair comparison across methods, all algorithms are initialized with the same random seed ($42$) and the identical initial dataset $D$. Additionally, all reported results are totally replicable and traceable, as the random factors are under control.

All the configurations included in the full configuration space are as follows:\\
$r\in R=\{0,0.1,0.2,0.3,0.4,0.5,0.6,0.7,0.8,0.9,1\}$,\\
$n\in N=\{4,6,8,10,12,24,48,60\}$,\\
$s\in S=\{1,2,3,4,5,6,7,8,9,10\}$, especially $s\le8$ when $n=60$.

Altogether $858$ configuration values in total.

\subsection{Material datasets}

This subsection introduces the two datasets of our work. The materials are sampled from the two datasets with a given ratio.

\subsubsection{Materials Project}

The Materials Project (MP)~\cite{jain2013commentary} is a large-scale, open-access repository of computed inorganic materials properties generated primarily via high-throughput \textit{ab initio} density functional theory workflows. Crystal structures are sourced from experimental databases and hypothetical prototypes, then standardized through space-group symmetrization and conventional cell selection before property evaluation following well-documented protocols adopting generalized gradient approximation. The Materials Project provides a semantically rich schema that links each material identifier to full provenance covering input structures, calculation parameters and code versions, computed thermodynamics including formation energies and phase stability from convex hull analyses, electronic structure information for band structures and density of states, mechanical and transport descriptors, and defect and surface models for selected systems. Data are accessible through a versioned REST API and Python client tools, with ontology-aligned fields capturing synthesis relevant information including energy above hull and decomposition pathways, while computed entries support reproducible and queryable workflows across consistent calculation sets.

\subsubsection{Joint Automated Repository for Various Integrated Simulations}

Joint Automated Repository for Various Integrated Simulations~\cite{choudhary2020joint} is a comprehensive material data ecosystem combining first principles calculations classical force field simulations and machine learning surrogates to deliver standardized property benchmarks across two dimensional bulk and molecular systems. \textit{ab initio} data largely derive from DFT calculations for layered materials and PBE methods~\cite{madsen2007functional} paired with spin orbit coupling in applicable cases while further supplemented by many body perturbation related workflows tight binding calculations and phonon analyses to support optoelectronic mechanical and vibrational property research. Semantically JARVIS organizes all research content as defined tasks with clear metadata recording relevant computational details exchange correlation functional selections k point mesh arrangements pseudopotential configurations convergence standards and post processing procedures. It supplies abundant characteristic parameters including exfoliation energies dielectric tensors elastic constants effective masses and topological invariants through stable unique identifiers and dedicated application programming interfaces. The framework prioritizes standardized benchmark evaluations and guaranteed result reproducibility facilitating direct model comparisons and clear traceability across different computational approaches.

\subsection{Property prediction tasks}

This subsection introduces the six prediction tasks used in AutoMatBench.

\subsubsection{Band gap}

Band gap~\cite{jain2013commentary,ong2015materials,zhuo2018predicting} task is defined as a regression problem to predict the electronic band gap in eV of inorganic crystalline materials. The dataset's inputs are crystal structures referenced to Materials Project entries along with structure derived descriptors while the target label is a continuous band gap value computed through standard DFT workflows adopted by the Materials Project following conventional GGA or PBE approaches that tend to produce values lower than experimental measurements. Labels reflect the fundamental gap obtained from calculated electronic band structure information. Metallic systems are assigned values close to zero and numerical thresholds support metallic classification in preprocessing workflows. Detailed metadata records original material identifiers and calculation configurations from the Materials Project ensuring full reproducibility and objective evaluation for models trained directly on structural information or predefined features.

\subsubsection{Bulk modulus}

Bulk modulus~\cite{jain2013commentary,ong2015materials,de2015charting} task aims to predict the bulk modulus in GPa for inorganic crystals reflecting inherent resistance toward volumetric compression. Input content mainly adopts crystal structure information and can integrate feature descriptions generated from component composition and structural characteristics. The corresponding label maintains continuous numerical form obtained from energy volume equation of state results calculated through standard elasticity workflows based on density functional theory. All data entries follow unified unit standards and consistent computational settings while relevant uncertainty information exists in implicit form without direct display. Fixed grouping modes support stable benchmark verification during model evaluation and every sample connects with exclusive material identification content as well as original elastic tensor data used for calculating bulk modulus values.

\subsubsection{Formation energy}

Formation energy~\cite{jain2013commentary,ong2015materials,castelli2012new} task targets the formation energy per atom in eV per atom a thermodynamic descriptor indicating stability relative to elemental references. Inputs are the crystalline structures and associated compositions mapped to material ids. Labels are computed formation energies from high throughput DFT pipelines using consistent reference states and applicable corrections throughout calculation processes. These values can combine with compositional energies to derive energy above hull while the label defined for this task remains strictly the per atom formation energy. Negative values represent thermodynamic stabilization relative to pure elements and near zero or positive values reflect metastability or instability of corresponding materials.

\subsubsection{Metallicity}

Metallicity~\cite{zhuo2018predicting} is a binary classification task that predicts whether a material holds metallic or nonmetallic properties including semiconductor and insulator features based on electronic structure information. In this research, it is treated similar to a regression problem. Inputs adopt crystal structures linked to the Materials Project and relevant derived descriptors. The label adopts categorical binary values determined from calculated band gap data. Materials are recognized as metallic when the DFT calculated gap stays within a small threshold range while other substances fall into the nonmetallic category. The dataset may contain imbalanced sample distributions and official preset data splits maintain stable evaluation standards throughout testing processes. The judgment standard reflects inherent limitations existing in conventional GGA and PBE band gap calculation methods. Materials with narrow gap values near the classification boundary show sensitivity to numerical details during computation which carries importance for modeling research focusing on uncertainty analysis.

\subsubsection{Refractive index}

Refractive index~\cite{jain2013commentary,ong2015materials,petousis2017high} task predicts the static refractive index of inorganic materials with no assigned measurement units. For simplification, no directional refraction index is considered. Inputs follow standard structure based representations connected to Materials Project references. The target label corresponds to refractive index values acquired from first principles linear response or density functional perturbation theory workflows relying on dielectric tensor data collected at zero frequency. Results are condensed into a single representative scalar determined from orientational averages of electronic contributions. The property carries no standard measurements and inherent anisotropy gets simplified into one unified value to suit benchmarking needs while explicit uncertainty information is not commonly provided within recorded data.

\subsubsection{Shear modulus}

Shear modulus~\cite{jain2013commentary,ong2015materials,de2015charting} task focuses on predicting the shear modulus in standard GPa units which characterizes material resistance toward shape change while maintaining constant volume. Inputs adopt crystal structure data referenced to the Materials Project and supporting features cover symmetry density and statistical information derived from chemical composition. The label presents continuous shear modulus values calculated from DFT elastic tensors following Voigt averaging standard~\cite{voigt1908lehrbuch} or else. The dataset clearly defines the adopted averaging approach and maintains unified unit settings across all entries.

\subsection{Evaluated ML models}

This subsection introduces five ML models evaluated with AutoMatBench in our experiments.

\subsubsection{ALIGNN}

Atomistic Line Graph Neural Network (ALIGNN)~\cite{choudhary2021atomistic} is a graph neural network tailored for crystalline materials that emphasizes explicit encoding of many body interactions through a dual graph message passing scheme built with an atomistic graph and an auxiliary line graph for representing three body bond angle relationships. Atoms serve as nodes with learnable embeddings initialized from elemental attributes. Edges record interatomic distances and bond angle contexts processed with radial basis function expansions. Its core architecture applies alternating edge gated graph convolution operations on the two graphs to capture key coordination patterns closely connected to material properties. A global pooling module aggregates node feature states to support scalar property prediction tasks. The model delivers strong performance for formation energy, band gap, elastic modulus, and stability classification applications where structure related local atomic environments dominate material behaviors and maintains competitive results on mainstream regression benchmarks and standard evaluation leaderboards for material research datasets.

\subsubsection{SchNet}

Sch\"utt's Network (SchNet)~\cite{schutt2017schnet,schutt2018schnet}, a variant of Deep Tensor Neural Networks (DTNN)~\cite{schutt2017quantum}, is a continuous filter convolutional neural network designed for molecular and crystalline systems that operates directly on nuclear charge information and three dimensional atomic positions. Interatomic interactions are modeled through continuous filter convolution modules defined with radial basis expansions of pairwise atomic distances within a preset cutoff range to build smooth and differentiable potential descriptions. The core structure alternates continuous filter interaction blocks and independent atom feature updates while sum pooling operations maintain consistent representation for extensive material properties. SchNet achieves reliable performance in quantum chemical property prediction covering energy force and dipole related outputs and delivers strong generalization for crystalline property regression tasks.

\subsubsection{CrabNet}

Compositionally restricted attention based network (CrabNet)~\cite{wang2021compositionally} is a composition only neural architecture built for material systems without available crystal structure data. It applies self attention mechanisms across element tokens through learned representation built from compositional information and adopts multi head Transformer modules to capture interaction patterns among constituent elements. The model operates without explicit structural input features and learns underlying connections between chemical composition and material properties while generating corresponding uncertainty estimation supported by residual connection design and dropout regularization strategies. It performs effectively on prediction tasks dominated by compositional influencing factors covering approximate formation energy evaluation, elastic constant, band gap, and preliminary screening for hardness and critical temperature values when only chemical formula content can be accessed. In this research, CrabNet is also used in structure-known tasks for consistent comparison.

\subsubsection{RF-SCM}

RF-SCM~\cite{dunn2020benchmarking} integrates the Random Forest regressor~\cite{kovacs2021lsst} and the Sine Coulomb Matrix~\cite{faber2015crystal} while adopting feature sets derived from Magpie descriptors~\cite{ward2016general}. The applied features cover stoichiometric statistical values average and variance information of elemental properties alongside basic heuristic rules that reflect periodic trends across chemical elements. The model builds on ensemble decision tree structures supported by bootstrap aggregation strategies achieving stable performance with limited hyperparameter adjustment and clear interpretable feedback on feature contribution levels. RF-SCM shows reliable adaption for datasets with limited sample size or obvious noise and performs well on tasks dominated by compositional influences. It supports preliminary evaluation for formation energy analysis, band gap, metallicity, and mechanical tendency recognition. Similar to CrabNet~\cite{wang2021compositionally}, it also works when crystal structure data cannot be obtained.

\subsubsection{MEGNet}

Materials Graph Network~\cite{chen2019graph} is a unified graph neural network that combines atom bond and global state updates through edge node state message passing mechanisms. Nodes represent atoms with elemental embeddings while edges capture pairwise interactions defined by distance and radial basis features. A global state vector records core conditions such as temperature pressure and domain related labels. Each block conducts sequential feature updates with skip connection structures followed by readout pooling modules for final property prediction. Trained on large scale DFT datasets the model reaches high accuracy in formation energy prediction and can adapt to band gap, elastic modulus, dielectric constant, and other scalar or tensor property tasks.

\section{Experiment}


In this section, we present four comprehensive evaluations of AutoMatBench. These experiments are designed to validate the effectiveness of this framework in identifying optimal models and configurations while reducing computational overhead with more than a half. The outcomes demonstrate that AutoMatBench can recover basic conclusions from exhaustive MatBench evaluations and OOD studies within just $12$ optimization steps, meanwhile achieving a high cost reduction. Based on our results, the majority of suggested configuration settings is $n\le24$ and $s=7$ or $8$ (except band gap~\cite{jain2013commentary,ong2015materials,de2015charting} task, the best configuration is $s=1$). The ratio $r$ is not significant.

\subsection{Experimental Setup}

All the experiments are conducted on a server equipped with an Intel Xeon CPU E5-2620 v3 of x86 architecture, with 63GB RAM and Ubuntu 20.04.6 LTS. The software environment was built using Anaconda with Python 3.7$\sim$3.12, with deep learning models implemented in \texttt{PyTorch} and \\\texttt{TensorFlow}. Data sources include the MP dataset~\cite{jain2013commentary} with \texttt{mp$\_$api.client.MPRester} and the JARVIS dataset~\cite{choudhary2020joint} with \texttt{jarvis.db.figshare.data}.

\subsection{Experiment set 1: Searching for the ML model that has the best prediction performance}

This experiment aims to identify the top-performing AI model within the AutoMatBench search space. This problem is the basic benchmarking aim for AI model researchers. It has a variant form, that is, to search for the configuration ($r,n,s$) in the entire space where a certain model achieves the best performance. For developers of AI models, this problem is significant for submitting outstanding performances on a leaderboard that has less restrictions. Our experiment is done on the variant form.

We evaluated five models on the band gap task~\cite{jain2013commentary,ong2015materials,zhuo2018predicting} and the RF-SCM model~\cite{dunn2020benchmarking} on all six tasks.

\begin{figure*}[ht]
\centering
\includegraphics[scale=0.32]{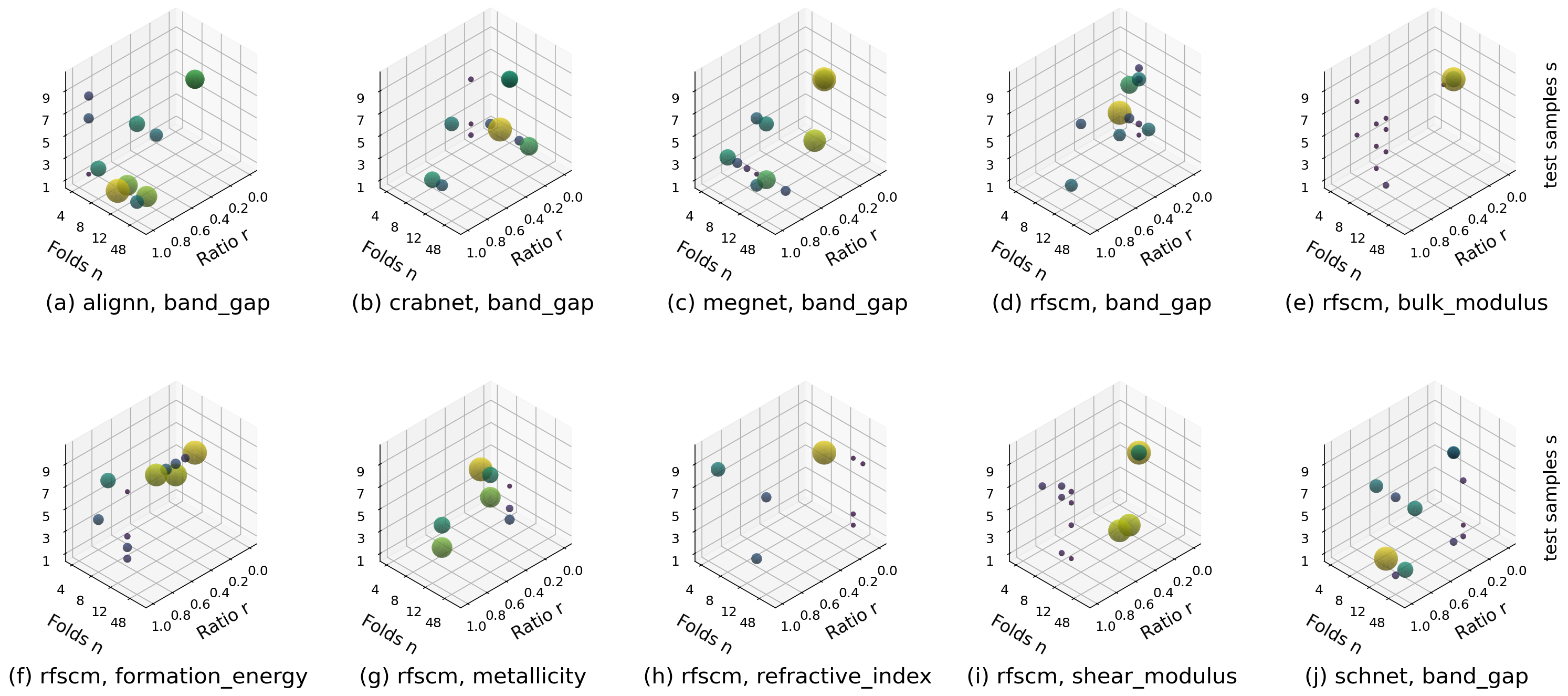}
\caption{This figure reports the wide discrepancies among evaluated MAEs while taking different configurations across different models and tasks. Sphere color/scale: MAE. Position: config.}\label{fig:spheres}
\end{figure*}

The first observation is that, in the configuration of hundreds of configurations, the performance of an AI model varies considerably. The maximum discrepancy is $10.045$ times (max $8.899$, min $0.886$), for the RF-SCM model~\cite{dunn2020benchmarking} on the refractive index prediction task~\cite{jain2013commentary,ong2015materials,petousis2017high}. Figure~\ref{fig:spheres} shows the wide range of multiple models' MAEs on multiple tasks. The color and the scale of spheres correspond to the MAE, and the position in the three-dimensional space corresponds to the configuration ($r,n,s$). Note that we excluded some outliers in the bulk modulus task~\cite{jain2013commentary,ong2015materials,de2015charting}. The reason is that some materials have extreme large bulk modulus ($>10,000$), and this may cause divergence in the training process.

The second observation is, for all tasks and models, the Bayesian optimizer finds an acceptable configuration within mostly $12$ steps. In most conditions, the optimal $s$ value is less than or equal to $7$. This indicates no more than $7$ test samples are required. The $n$ values varies, but for most time, taking less than $24$ is feasible. This eliminates at least half of benchmarking cost compared to Omee et al.'s study~\cite{omee2024structure} without reliability loss. Figure~\ref{fig:1st_1} and~\ref{fig:1st_2} show the process and results of optimization in this experiment. Red points stand for newly evaluated MAE. Blue points stand for the minimum. The yellow circle indicates the optimal configuration and its outcome.

\subsection{Experiment set 2: Searching for the benchmark configuration that has the maximum degree of discrimination}

This experiment aims to identify the configuration ($r,n,s$) in the entire space that discriminates the performances of models. An adequate configuration may reveal the fundamental advantages and disadvantages of certain models. This problem is significant for benchmark leaderboard designers and researchers.

We evaluated RF-SCM~\cite{dunn2020benchmarking} and CrabNet~\cite{wang2021compositionally} models on all six tasks.

Our observation is, in the six tasks, more than a half of tasks' optimal discrimination configuration takes low $s$ quantity ($s\le2$). The $s$ value stands for the test samples. As a result, taking a few samples affects the stability of performance results, for models may occasionally get mistaken on small datasets. Therefore, a bigger $s$ is necessary. Besides $s$ value, the $r$ ratio has little influence on results. As for the number of folds $n$, the range from $4$ to $8$ is enough. Figure~\ref{fig:1dual} shows the process and results of optimization in this experiment. This eliminates about 75\% benchmarking cost compared to Omee et al.'s study~\cite{omee2024structure} without reliability loss. Red points stand for newly evaluated MAE. Blue points stand for the minimum. The yellow circle indicates the optimal configuration and its outcome. However, for stability of evaluation, the proper $s$ value should be set as $7$ or $8$ for tasks (a), (d) and (g). The experiment result for $s=7$ or $s=8$ also indicates an enough sufficiency with very little discrimination ability loss. As for task (f), the evaluation fails. The possible explanation is that the prediction for shear modulus~\cite{jain2013commentary,ong2015materials,de2015charting} fails to converge, as models encounter extreme large values.

\begin{figure*}[ht]
\centering
\includegraphics[scale=0.25]{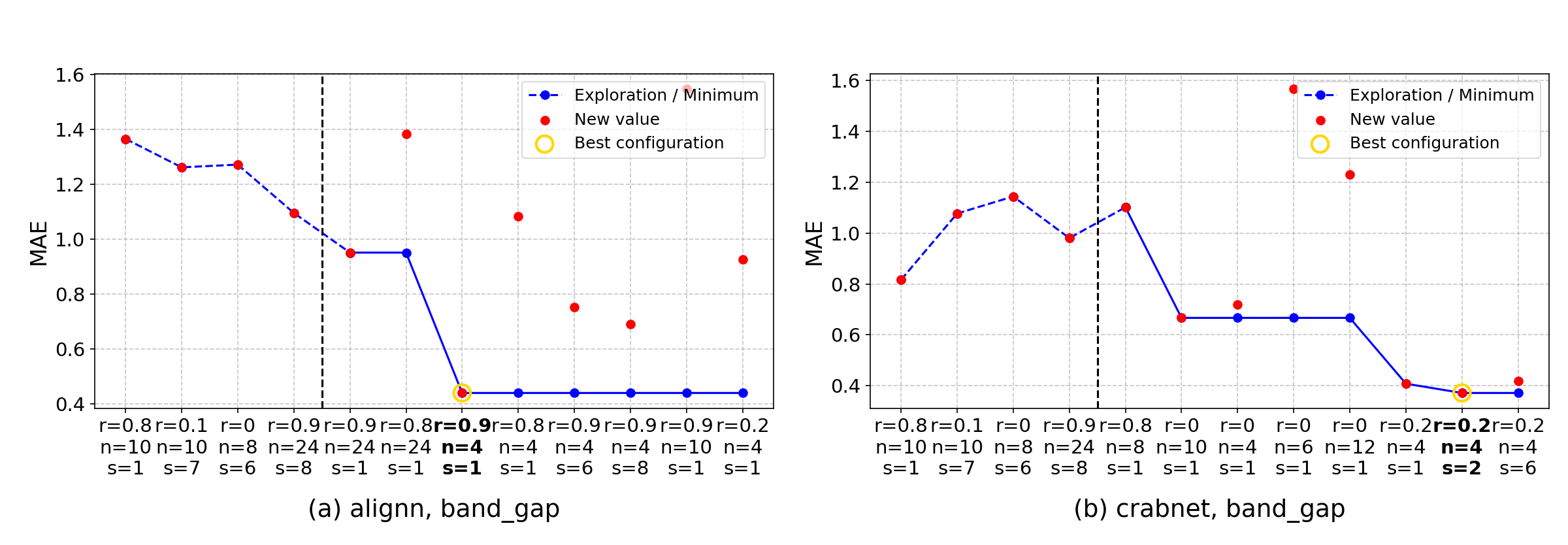}
\caption{This figure shows two examples [(a) and (b)] of optimization in Experiment 1. The optimal configurations are found within 12 steps. The full results are presented in the appendix.}\label{fig:1st_demo}
\end{figure*}

\begin{figure*}[ht]
\centering
\includegraphics[scale=0.25]{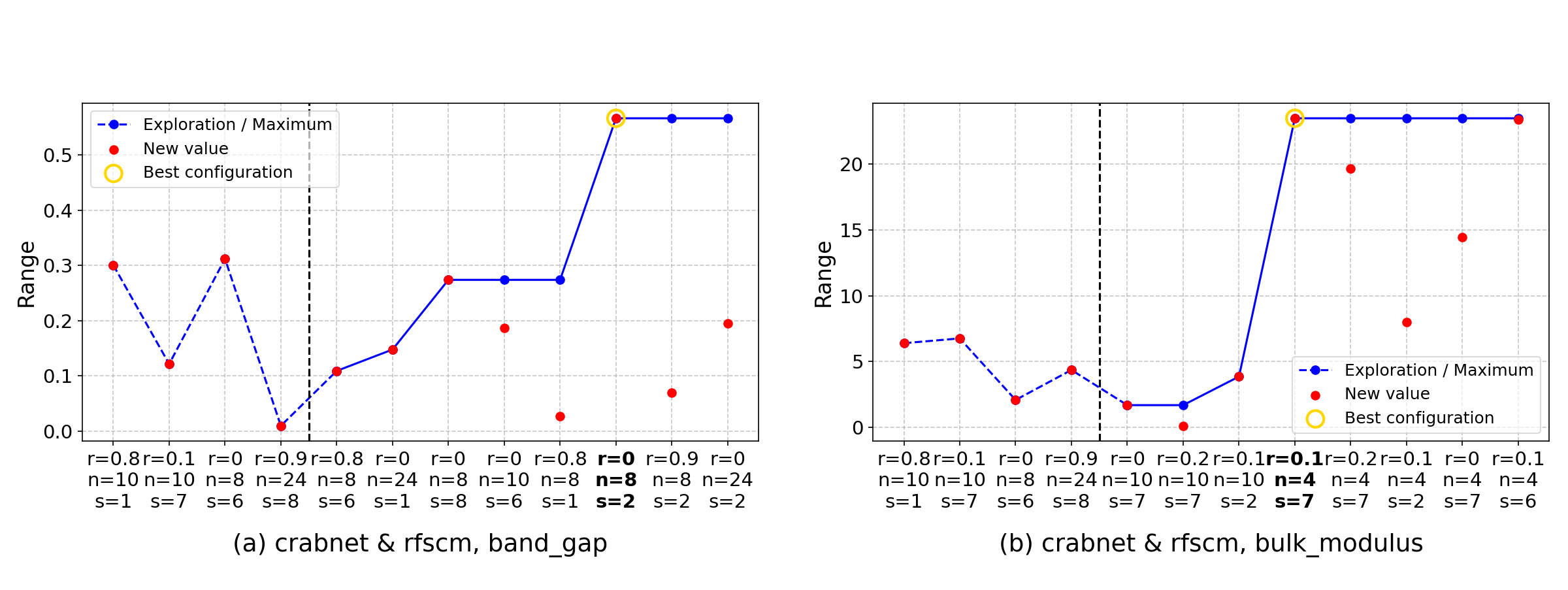}
\caption{This figure shows two examples [(a) and (b)] of optimization in Experiment 2. The optimal configurations are found within 12 steps. The full results are presented in the appendix.}\label{fig:1dual_demo}
\end{figure*}

\begin{figure*}[ht]
\centering
\includegraphics[scale=0.25]{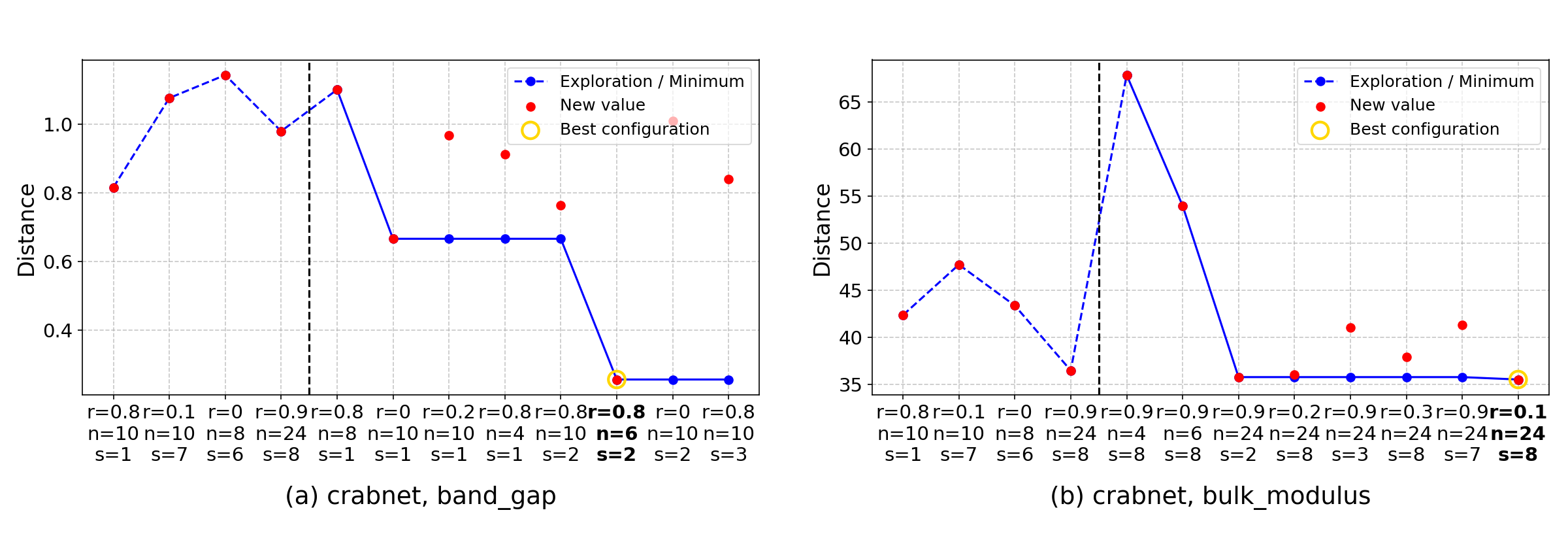}
\caption{This figure shows two examples [(a) and (b)] of optimization in Experiment 3. The optimal configurations are found within 12 steps. The full results are presented in the appendix.}\label{fig:2nd_demo}
\end{figure*}

\subsection{Experiment set 3: Searching for the benchmark configuration closest to the ground-truth of model prediction performance}

This experiment aims to identify the configuration ($r,n,s$) in the entire space that yields the performances closest to the ground-truth. The ground-truth MAEs of models are defined as the suggested configuration $(r,n,s)=(0,48,10)$ by the existing study~\cite{omee2024structure}. This assumption ensure that main conclusions with AutoMatBench is capable of reproducing existing findings. A configuration that yields results seriously inconsistent with the ground-truth is considered as unreliable.

We evaluated RF-SCM~\cite{dunn2020benchmarking} and CrabNet~\cite{wang2021compositionally} models on all six tasks.

Figure~\ref{fig:2nd_1} and~\ref{fig:2nd_2} show the process and results of optimization in this experiment. For the ratio parameter $r$, no significant influence is found. For the number of folds $n$, every experiment outcome reports that $s\le24$ is enough. This eliminates at least $50\%$ benchmarking cost compared to Omee et al.'s study~\cite{omee2024structure} without reliability loss. As for parameter $s$, most of tasks takes $s=7$ or $s=8$. The only exceptions are (a), (g), and (i) with low $s$ values. However, for task (i), setting $s=7$ instead of $1$ is feasible. Therefore, for tasks except band gap~\cite{jain2013commentary,ong2015materials,zhuo2018predicting}, a bigger $s$ is necessary. Red points stand for newly evaluated MAE. Blue points stand for the minimum. The yellow circle indicates the optimal configuration and its outcome.

\subsection{Experiment set 4: Case study on difference between approximate data and real dataset}

Finally, in case of inappropriate results using the synthesized data, we conducted a case study comparing models trained on the approximate data computed with JARVIS dataset~\cite{choudhary2020joint} versus real data from MP dataset~\cite{jain2013commentary}. The tasks taken are band gap~\cite{jain2013commentary,ong2015materials,zhuo2018predicting} and refractive index\cite{jain2013commentary,ong2015materials,petousis2017high} prediction. All five models are evaluated.

The first data approximation method is taking the band gap value ($\le$ 0.001 eV) instead of metallicity. The principle of this method is based on the difference in band gap characteristics between metals and nonmetals: metals have no obvious band gap or an extremely small band gap, which enables electrons to move freely and show metallic properties, while nonmetals have a significant band gap. This approximate method is applicable to the preliminary screening of materials, where it can effectively distinguish materials with metallic properties from insulators by means of the band gap criterion. This approach may fail on metalloids.

Another data approximation method for calculating the refractive index is based on the dielectric constants (mepsx, mepsy, and mepsz), using the formula $n=\sqrt{\varepsilon}$. The principle of this method lies in the relationship between the refractive index and the relative permittivity: for most non-magnetic materials, the relative permeability is approximately $1$, so the refractive index can be approximated as the square root of the relative permittivity, which describes the material's response to an electric field. This approach may fail when materials are non-magnetic~\cite{wangberg2006nonmagnetic}. It may also fail when the frequency dependence of the dielectric constant and the anisotropy of the material must be considered.

The evaluation results are shown in Table~\ref{tab:exp4}. For any AI model and each task, the discrepancy between MAE on the approximate dataset and MAE on the real dataset is within $5.4\%$ times. This outcome is an acceptable error. Therefore, we conclude that no significant influence on using the approximate datasets exist. This experiment justifies the soundness of this research.

\begin{table}[ht]
\caption{The results of the case study. This table reports that approximation adopted in our study is reliable.}
\label{tab:exp4}
\begin{tabular}{llccc}
\hline
Task              & Model   & Approx      & Real    & Diff               \\\hline
Band gap          & ALIGNN  & 0.4811      & 0.4778  & 0.33\%$\uparrow$   \\
Band gap          & SchNet  & 0.7149      & 0.7584  & 4.35\%$\downarrow$ \\
Band gap          & CrabNet & 0.5400      & 0.5401  & 0.01\%$\downarrow$ \\
Band gap          & RF-SCM  & 0.4016      & 0.3895  & 1.21\%$\uparrow$   \\
Band gap          & MEGNet  & 0.6773      & 0.6233  & 5.40\%$\uparrow$   \\
Refractive index  & ALIGNN  & 0.1993      & 0.1938  & 0.54\%$\uparrow$   \\
Refractive index  & SchNet  & 0.3035      & 0.2760  & 2.76\%$\uparrow$   \\
Refractive index  & CrabNet & 0.2298      & 0.2459  & 3.38\%$\downarrow$ \\
Refractive index  & RF-SCM  & 0.2219      & 0.2263  & 0.44\%$\downarrow$ \\
Refractive index  & MEGNet  & 0.2281      & 0.2304  & 0.24\%$\downarrow$ \\\hline
\end{tabular}
\end{table}

\section{Limitation}

Despite the promising efficacy and practical insights delivered by AutoMatBench in both ID and OOD benchmarking, some limitations remain to be addressed.

The current evaluation scope is constrained to six representative MatBench tasks and five mainstream AI models. It excludes a broader spectrum of advanced prediction tasks in MatBench, such as emerging large-scale material screening models.

Additionally, the OOD construction paradigm adopted in this work inherits clustering-based data partitioning from prior studies, which only covers structural and compositional distribution shifts while ignoring other realistic OOD possibility. The current tool supports neither multi-modal data fusion nor cross-domain migration evaluation across inorganic crystals, polymers, and amorphous materials. This is still distant from the original intention of AI for materials -- accelerating the discovery of novel and excellent materials.

The future work is threefold. First, a larger scale of benchmarking on more models and more tasks. Second, attributing concrete causes of evaluated results to inner structures of AI prediction models. Finally, extension of AutoMatBench for the support of multi-modal data fusion and cross-domain evaluation across diverse material systems for better material discovery.

\section{Conclusion}

This study proposes AutoMatBench, an automatic toolkit integrated with Bayesian optimization. Experimental results show that, in all tasks, within twelve optimization steps, AutoMatBench can identify target-compliant configurations for all involved tasks and AI models, which aligns with existing research with no big gaps and reduces over half of the experimental cost.

Based on our experimental results and comprehensive statistical analysis across all evaluated tasks, the majority of the optimal suggested configuration settings converge to a regime with the system size parameter satisfying $n\le24$ and the hyperparameter $s$ fixed at either $7$ or $8$. Notably, this general trend holds consistently for most property prediction tasks investigated in this work. The only exception is identified for the band gap prediction task, the best-performing and most robust configuration uniquely adopts $s=1$. Furthermore, the ratio hyperparameter $r$ exhibits negligible influence on the final model performance, showing no significant difference between using MP or JARVIS dataset. Consequently, the tuning of $r$ can be largely deprioritized in practical configuration optimization for the proposed framework. Increasing the optimization steps beyond twelve may yield better configurations, yet the performance improvement is limited.

AutoMatBench provides an efficient tool with innovative insights in enriching relevant theoretical and empirical research in the AI for materials.

\bibliographystyle{cas-model2-names}

\bibliography{reference}

\newpage
\appendix

\section{Full results of Experiment set 1, 2, and 3}

This section presents the complete experimental results for Experiment sets 1, 2, and 3.

\begin{figure*}[ht]
\centering
\includegraphics[scale=0.25]{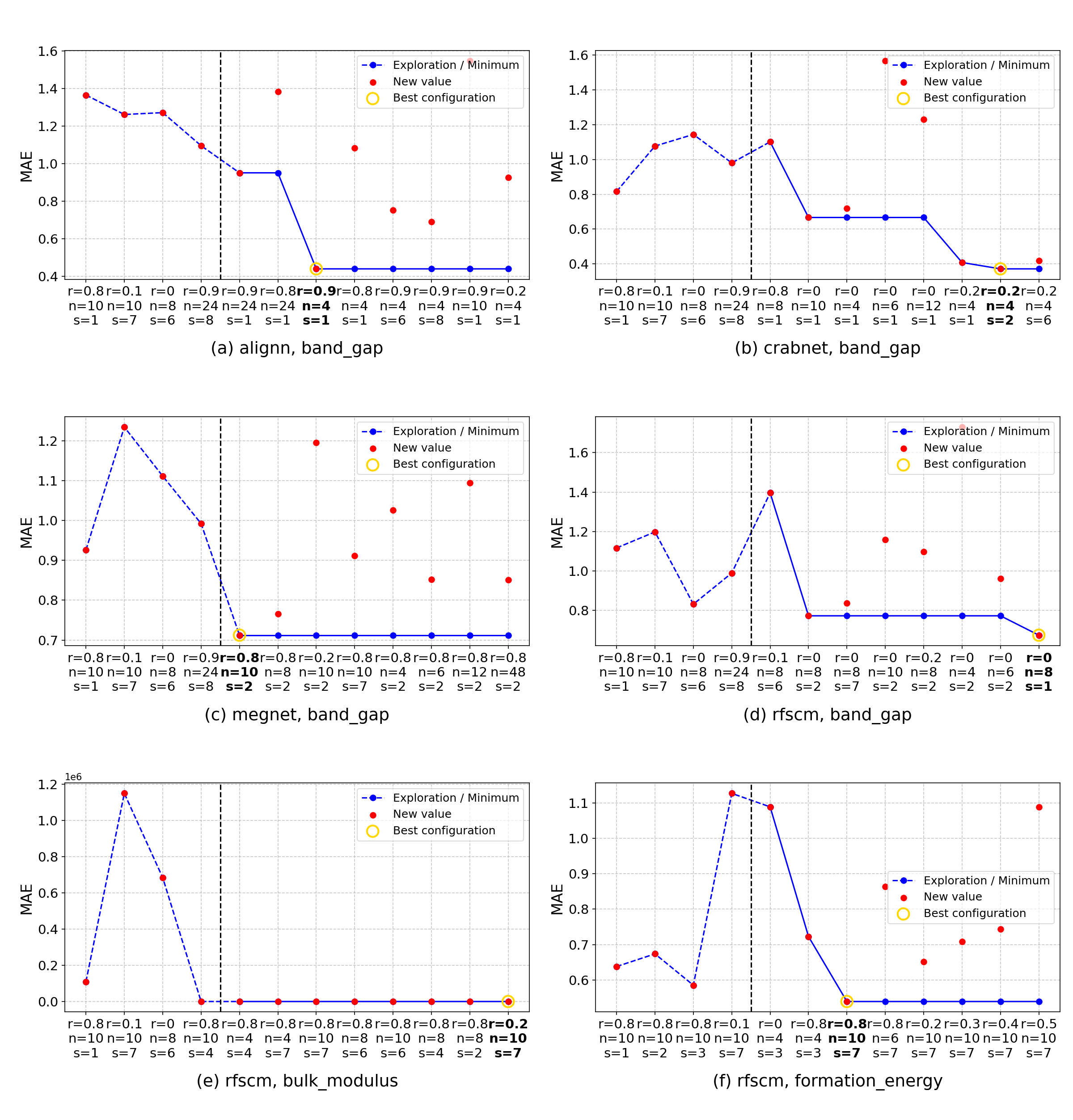}
\caption{This figure shows the entire processesand results of optimization in Experiment 1. (Part(a)$\sim$(f))}\label{fig:1st_1}
\end{figure*}

\begin{figure*}[ht]
\centering
\includegraphics[scale=0.25]{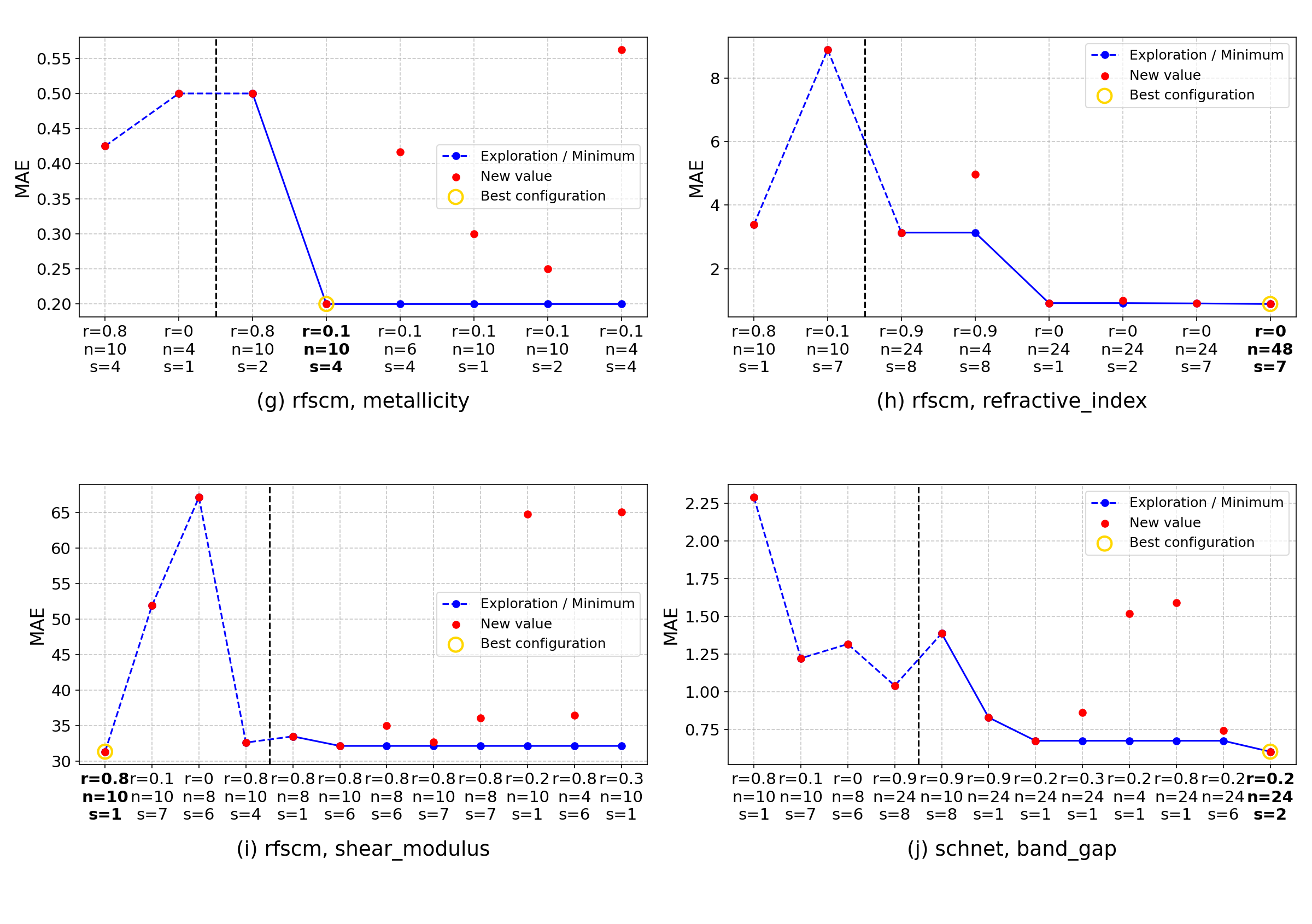}
\caption{This figure shows the entire processesand results of optimization in Experiment 1. (Part(g)$\sim$(j))}\label{fig:1st_2}
\end{figure*}

As illustrated in Figure~\ref{fig:1st_1} and Figure~\ref{fig:1st_2}, the detailed optimization processes and corresponding outcomes of Experiment 1 are clearly displayed across parts (a) to (j), covering the entire pipeline from initialization to final convergence.

The same full result data from Experiment sets 2 and 3 are also provided. Collectively, these results validate the effectiveness and stability of the proposed method across multiple experimental settings.

\begin{figure*}[ht]
\centering
\includegraphics[scale=0.38]{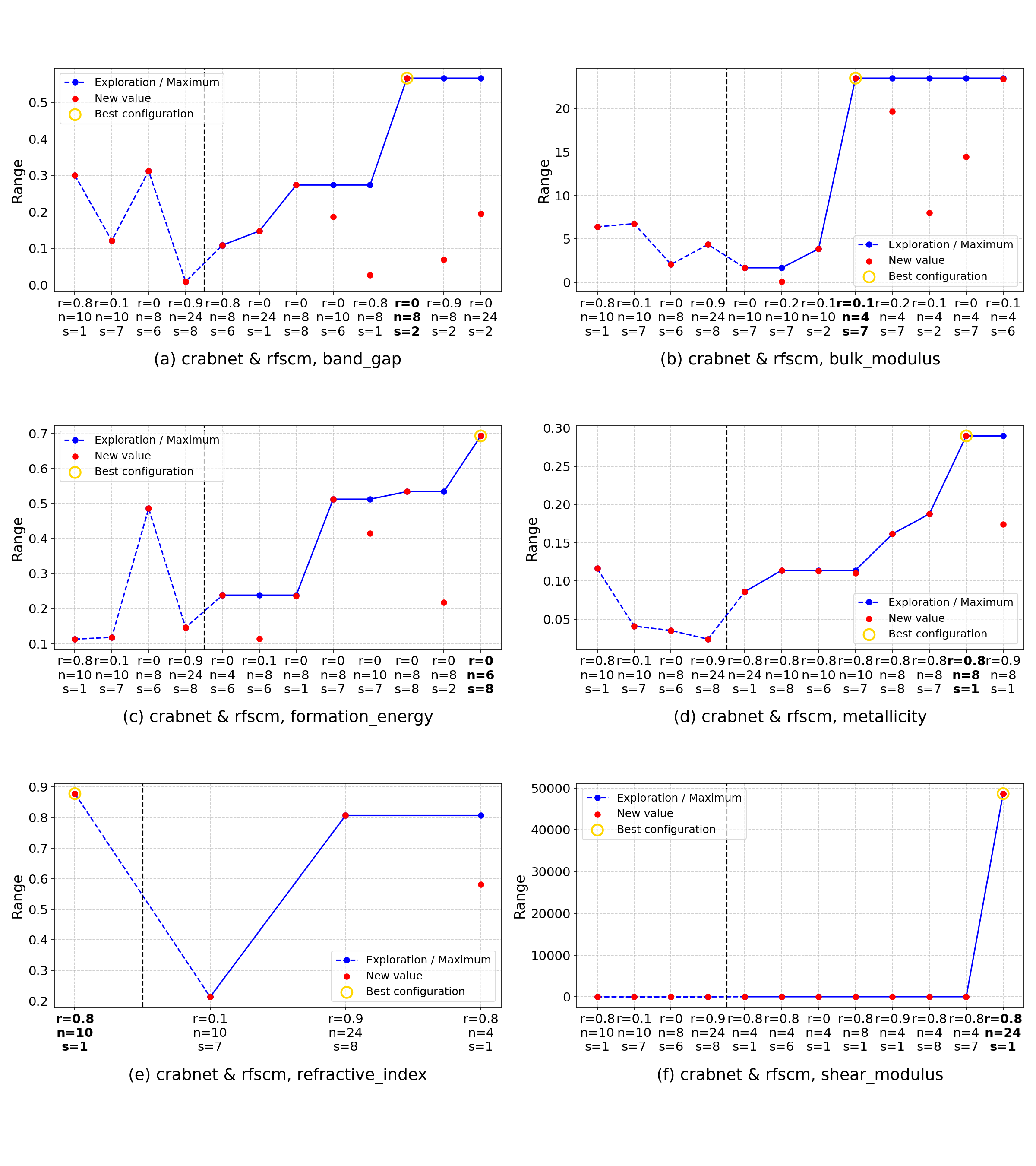}
\caption{This figure shows the entire processesand results of optimization in Experiment 2. Theoptimization aim is the maximize the performancegap between models. [Part (a)$\sim$(f)]}\label{fig:1dual}
\end{figure*}

\begin{figure*}[ht]
\centering
\includegraphics[scale=0.25]{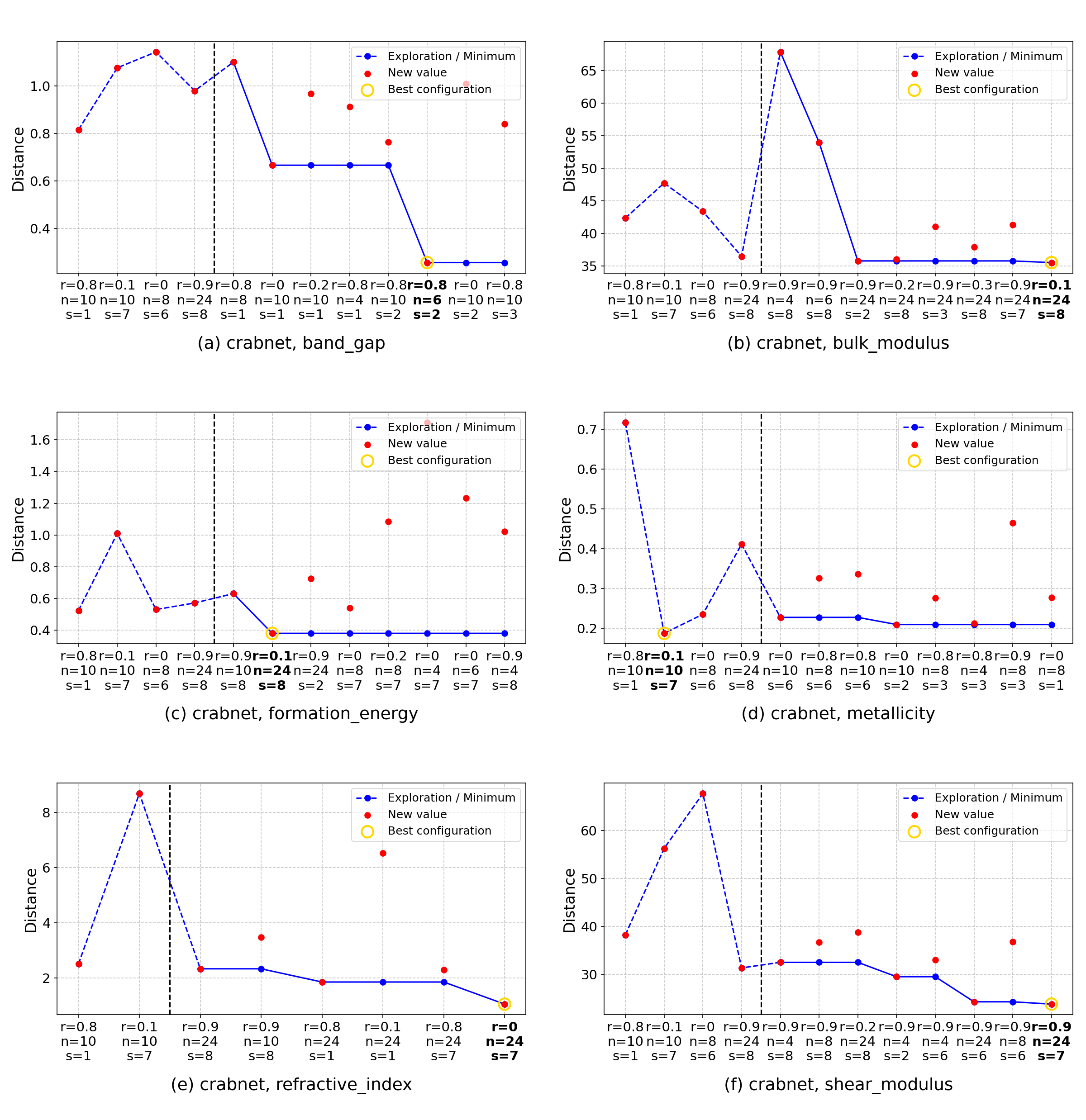}
\caption{This figure shows the entire processesand results of optimization in Experiment 3. Theoptimization aim is the minimize the performancedistance with the ground-truth results. [Part (a)$\sim$(f)]}\label{fig:2nd_1}
\end{figure*}

\begin{figure*}[ht]
\centering
\includegraphics[scale=0.25]{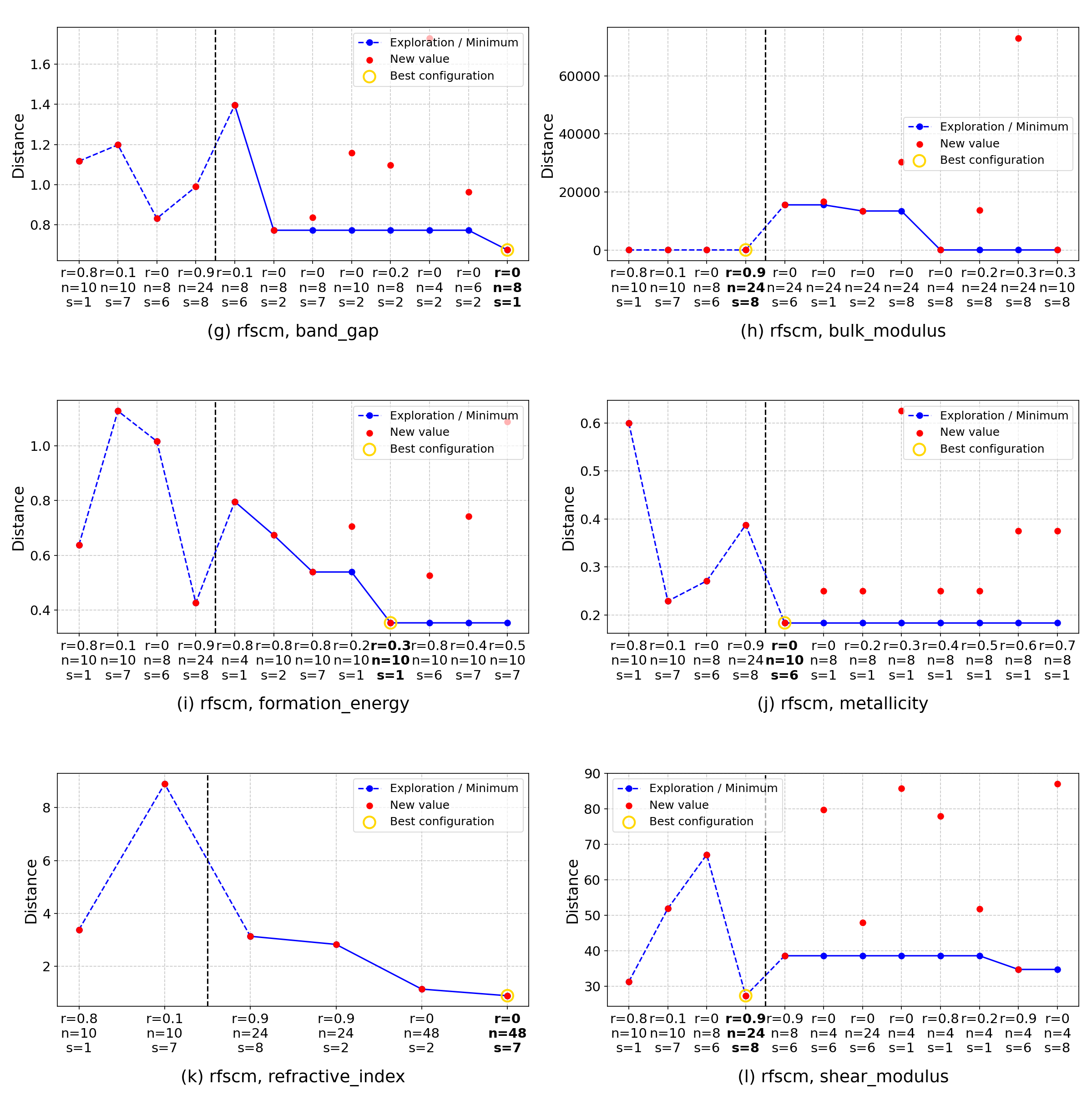}
\caption{This figure shows the entire processesand results of optimization in Experiment 3. Theoptimization aim is the minimize the performancedistance with the ground-truth results. [Part (g)$\sim$(l)]}\label{fig:2nd_2}
\end{figure*}

\end{document}